\documentclass{article}

\usepackage{PRIMEarxiv}

\usepackage[utf8]{inputenc} 
\usepackage[T1]{fontenc}    
\usepackage{hyperref}       
\usepackage{url}            
\usepackage{booktabs}       
\usepackage{amsfonts}       
\usepackage{nicefrac}       
\usepackage{microtype}      
\usepackage{lipsum}
\usepackage{graphicx}
\graphicspath{{media/}}     
\usepackage{footmisc}
\usepackage{url}
\usepackage{multirow}
\usepackage[noadjust]{cite}
\usepackage{amsmath,amsthm}
\usepackage{amssymb,amsfonts}
\usepackage{booktabs}
\usepackage{color}
\usepackage{ccaption}
\usepackage{booktabs}
\usepackage{float}
\usepackage{fancyhdr}
\usepackage{caption}
\usepackage{xcolor,stfloats}
\usepackage{comment}
\usepackage{caption}
\usepackage{graphicx}
\usepackage{float} 
\usepackage{bm} 
\usepackage{float}
\usepackage{subcaption}
\usepackage{ulem}
\graphicspath{{figures}}
\usepackage{cuted}  
\def \hv {{\bm{h}}}
\def \vv {{\bm{v}}}

\def \xv {{\bm{x}}}
\def \yv {{\bm{y}}}
\def \bv {{\bm{b}}}
\def \rv {{\bm{r}}}
\def \omegav {{\bm{\omega}}}

\def \Ecal {\mathcal{E}}
\def \Fcal {\mathcal{F}}
\def \Gcal {\mathcal{G}}
\def \Hbb {\mathcal{H}}

\def \Lcal {\mathcal{L}}

\def \Ncal {\mathcal{N}}

\def \Rcal {\mathcal{R}}

\def \Vcal {\mathcal{V}}

\def \Hbb {\mathbb{H}}

\def \Rbb {\mathbb{R}}

\title{Hyperbolic Hypergraph Neural Networks for Multi-Relational Knowledge Hypergraph Representation}

\author{
  Mengfan Li, Xuanhua Shi, Chenqi Qiao, Teng Zhang, Hai Jin \\
  National Engineering Research Center for Big Data Technology and System \\
  Services Computing Technology and System Lab, Cluster and Grid Computing Lab\\
  School of Computer Science and Technology\\
  Huazhong University of Science and Technology \\
  Wuhan, 430074, China\\
  \texttt{\{limf, xhshi,qiaocq, tengzhang, hjin\}@hust.edu.cn} \\
}

\begin{document}
\maketitle

\begin{abstract}
Knowledge hypergraphs generalize knowledge graphs using hyperedges to connect multiple entities and depict complicated relations. Existing methods either transform hyperedges into an easier-to-handle set of binary relations or view hyperedges as isolated and ignore their adjacencies. Both approaches have information loss and may potentially lead to the creation of sub-optimal models. To fix these issues, we propose the Hyperbolic Hypergraph Neural Network (H$^2$GNN), whose essential component is the \textit{hyper-star message passing}, a novel scheme motivated by a lossless expansion of hyperedges into hierarchies. It implements a direct embedding that consciously incorporates adjacent entities, hyper-relations, and entity position-aware information. As the name suggests, H$^2$GNN operates in the hyperbolic space, which is more adept at capturing the tree-like hierarchy.
We compare H$^2$GNN with $15$ baselines on knowledge hypergraphs, and it outperforms state-of-the-art approaches in both node classification and link prediction tasks.
\end{abstract}
\section{Introduction}

Knowledge hypergraphs are natural and straightforward extensions of knowledge graphs~\cite{chen2023poskhg, wang2023enhance,wu2023medical}. They encode high-order relations within diverse entities via hyper-relations and have been widely used in downstream tasks including question answering ~\cite{guo2021universal,zhang2020reinforcement}, recommendation system~\cite{yu2021self,tan2011using}, computer vision~\cite{li2023kbhn,DBLP:conf/aaai/ZengJBL23} and healthcare~\cite{wu2023megacare}. Generally, knowledge hypergraphs store factual knowledge as tuples (relation, entity$_1$, \dots, entity$_m$), where entities correspond to nodes and hyper-relations correspond to hyperedges.

The core of representation learning for knowledge hypergraphs lies in embedding hyperedges, which encompass multiple nodes and exhibit diverse types. Existing methods implement the direct approaches~\cite{wen2016representation,DBLP:conf/ijcai/FatemiTV020,DBLP:conf/acl/GuanJGWC20}. These methods commonly view hyperedges as isolated, independently learning embeddings without considering structural information with different hyperedges that share common nodes.
For instance, consider the tuple (education, Stephen Hawking, \textit{University College Oxford}, BA degree) and (locate, \textit{University College Oxford}, Oxford, England). Rather than learning in isolation, merging adjacent hyperedges based on the common node \textit{University College Oxford} can facilitate the deduction of an additional tuple (live in, Stephen Hawking, Oxford, England) within the \textit{live in} hyper-relation.
Therefore, skillfully integrating adjacencies becomes essential for capturing underlying relations, prompting us to delve into hypergraph neural networks, specialized in modeling structural information.

Existing hypergraph neural networks~\cite{feng2019hypergraph,yadati2019hypergcn,DBLP:conf/ijcai/HuangY21}, predominantly concentrate on a single hyper-relation, neglecting the abundant and diverse multiple hyper-relation types inherent in knowledge hypergraphs. Furthermore, they often overlook the importance of incorporating position-aware information.
Take \textit{(flight, Beijing, Shanghai, Guangzhou)} as an example, which means the flight takes off from Beijing and passes through Shanghai before landing in Guangzhou. According to clique expansion-based hypergraph models~\cite{feng2019hypergraph,yadati2019hypergcn}, it will be split into three distinct triples: \textit{(Beijing, flight, Shanghai)}, \textit{(Shanghai, flight, Guangzhou)}, and \textit{(Beijing, flight, Guangzhou)}, which lose the crucial information that Shanghai is an intermediate location and introduces unreal flight from Beijing to Guangzhou. 
Therefore, the order of entities in hyperedges, also known as position-aware information, holds semantic significance. A novel approach is imperative—one that considers the position of entities within hyperedges and accommodates multiple relations.

\begin{figure*}[htbp]
\centering
\scalebox{1.0}{
\includegraphics[width = 1.0\textwidth]{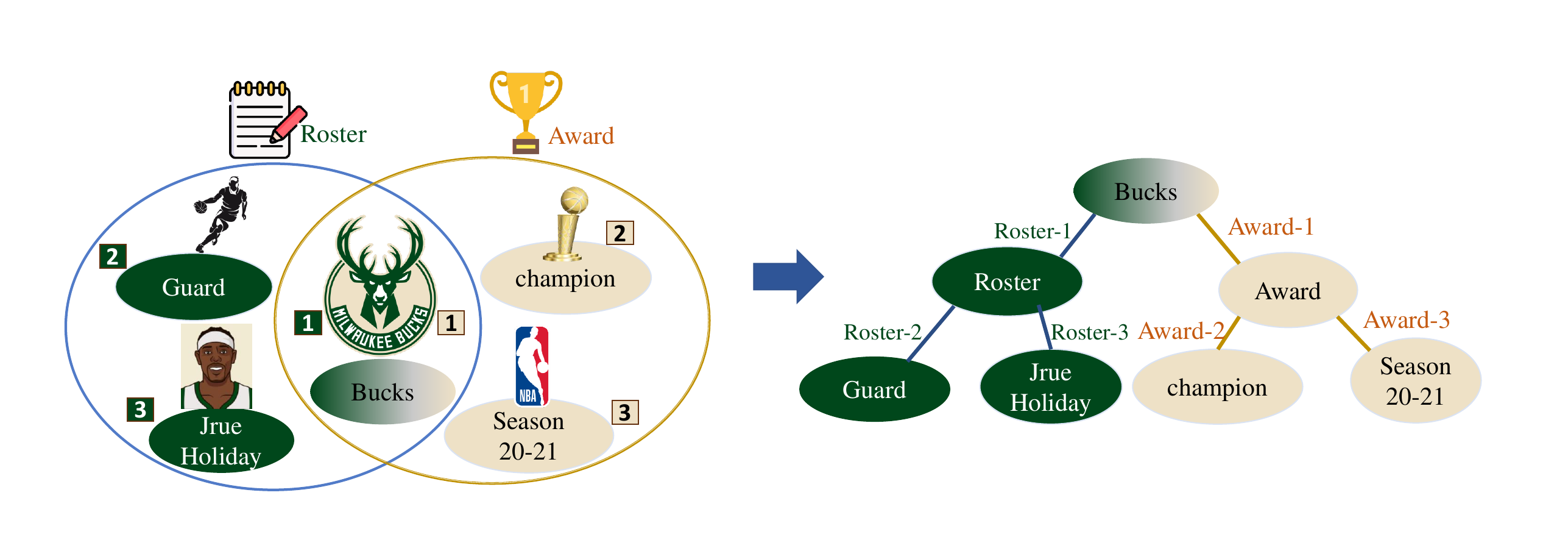}}
\vspace{-1em}
\caption{Considering the tuple \textit{(Roster, Bucks, Guard, Jrue Holiday)} and \textit{(Award, Bucks, champion, Season 20-21)}. We integrate position-aware information into the knowledge hypergraph by introducing relations like \texttt{Roster-1}, \texttt{Roster-2}, \texttt{Roster-3}, \texttt{Award-1}, \texttt{Award-2}, and \texttt{Award-3}. These relations correspond to their order within the tuple and hyperedges and the expansion results in a hierarchical structure.}
\label{hyper-star}
\end{figure*}

Based on these observations, we propose augmenting hypergraph neural networks, which are specifically crafted to comprehend hypergraph structures within knowledge hypergraphs, incorporating a specialized hyper-star message passing scheme tailored for knowledge hypergraphs. 
This scheme adeptly incorporates an extension to infuse both position-aware and adjacent information concurrently, resulting in the transformation of a knowledge hypergraph into a hierarchically structured, tree-like graph. 
Specifically, we introduce position-aware representations for each node, and then, in each HGNN layer, a two-stage message passing is performed, one is the aggregation of hyperedges embedding through the nodes they contain, while the other focuses on updating each node embedding by considering their positions, adjacent hyperedge embeddings, and hyper-relation embeddings. 

Unlike the approach presented in~\cite{DBLP:conf/ijcai/FatemiTV020}, which employs an embedding-based method for knowledge hypergraph presentation, our work pioneers consider the structure-based method
and extend GNNs to modeling knowledge hypergraphs, marking a novel and distinctive approach in this domain.
We notice that~\cite{fan2021heterogeneous} also utilizes GNN for learning knowledge hypergraph, however, it models hyperedge in a class-dependent way, that is for multiple type hyper-relations, they need to create a separate hypergraph for each type of hyper-relation, which cannot be satisfied in most situations. Specifically, each hypergraph is homogeneous. 
Instead, we view hyperedges as instance-dependent, modeling them based on actual instances directly with various hyperedge types and node types.
Furthermore, the hierarchy demonstrated in the message-passing process inspires us to explore a representation in hyperbolic space that can better capture the characteristics of scale-free and hierarchical graphs~\cite{shi2023ffhr,chami2019hyperbolic,muscoloni2017machine,chen2021fully}.

The contributions of this paper are as follows: 
\begin{itemize}
\item We put forward a new message-passing scheme catered specifically to hypergraphs, effectively consolidating nodes, hyper-relations, and entity positions within knowledge hypergraphs. This marks our initial endeavor to apply hypergraph neural networks to model multi-relational knowledge hypergraphs in an instance-dependent way. 
    \item We implement a versatile plug-and-play encoder H$^2$GNN, which can be easily concatenated with task-specific decoders and widely used in a wide range of downstream tasks. Experiments demonstrate the effectiveness of our proposed method.
\end{itemize}

\section{Preliminaries}
\noindent
In the realm of hypergraph neural networks, the endeavor to model multi-relational knowledge hypergraphs necessitates addressing novel challenges and opportunities. In this section, our focus is on addressing two fundamental questions.

\textit{Challenges: How to incorporate the potential position-aware information in multi-relational knowledge hypergraph?} 
For each tuple $(r,x_1,x_2, \dots, x_m)$ within the knowledge hypergraph, we create an interaction between the hyperedge and entities. To differentiate entities within the same hyperedge, we seamlessly introduce position-aware information based on the input order of the tuple. 
For instance, consider the tuple $(Roster, Bucks, Guard, Jrue$ $Holiday)$, which means \textit{The basketball player Jrue Holiday is a guard for the Milwaukee Bucks in the Roster}.
\textit{Roster} is the hyper-relation, representing the overall category of information. \textit{Bucks} is the first entity and specifies the team involved in the roster. \textit{Guard} is the second entity and provides additional details about the player's position in the team. \textit{Jrue Holiday} is the third entity and represents the specific player mentioned in the roster. 
To highlight the significance of position-aware information, we design an expansion to incorporate it into the hypergraph neural network on multi-relational knowledge hypergraph. Specifically, we construct additional relations \texttt{Roster-1}, \texttt{Roster-2}, \texttt{Roster-3} to establish the instantiated hyperedge \texttt{Roster} and its entities. 
As illustrated in Figure~\ref{hyper-star}, the hyperedges \texttt{Roster} and \texttt{Award} are connected with the nodes contained in the hyperedge. The newly generated relation is determined by the relation and the position of the node in the hyperedge. 
And if the hyper-relation contains more elements, for instance, a maximum of $n$ entities, then we initialize $n$ position-aware features to ensure compatibility with all existing examples within the dataset. 
This expansion transforms the hyperedge into a tree-like hierarchy. Subsequent message passing in the hypergraph neural network, aimed at capturing the hypergraph structure, operates on this expanded hierarchy, which now includes position-aware information.

\begin{figure}[!ht]
\begin{subfigure}{0.5\textwidth}
    \centering
    \includegraphics[scale=0.35]{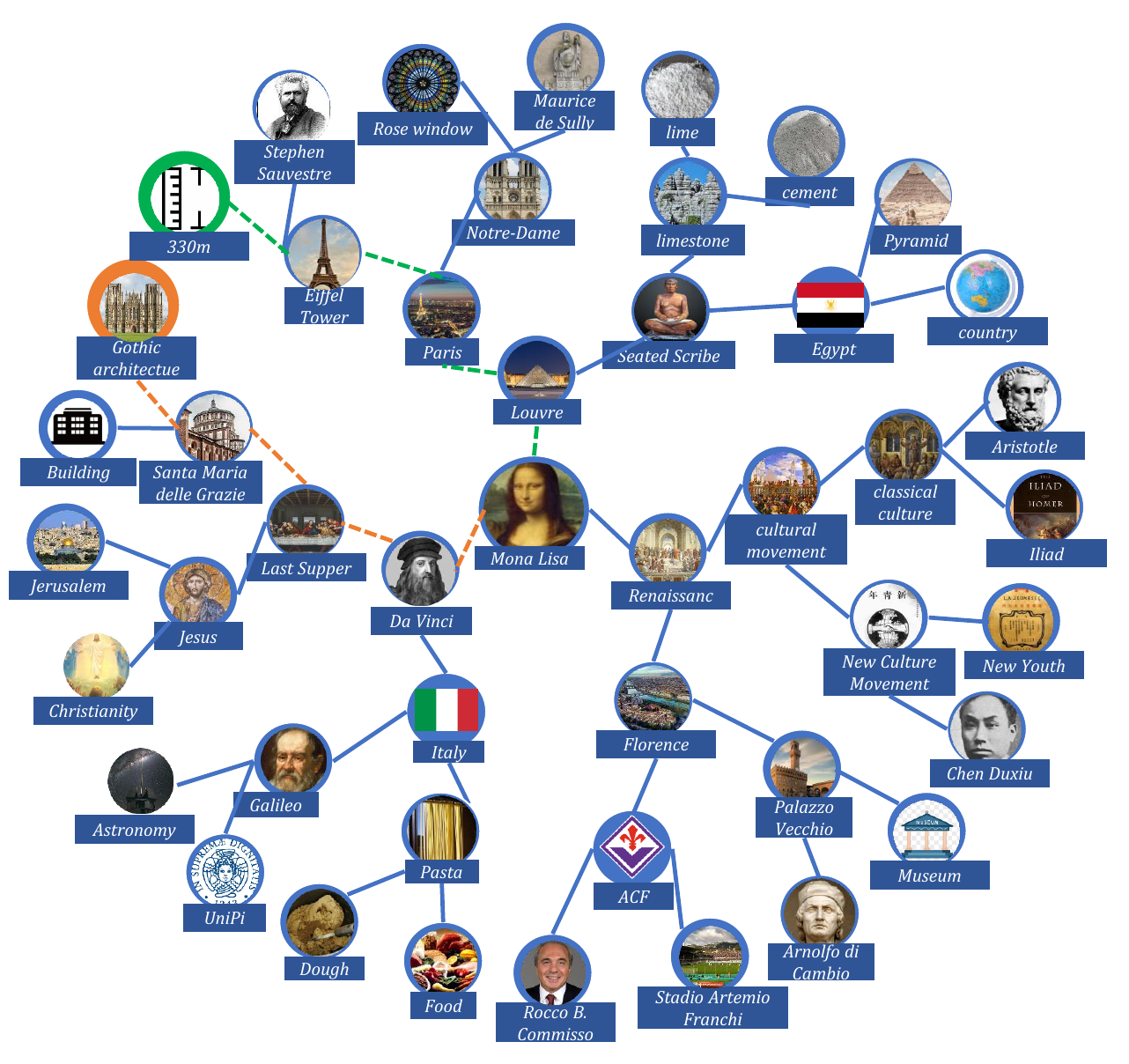}
    \caption{Hierarchy instantiation of \textit{Mona Lisa} and its related entities in Euclidean space.}
    \label{Eu}
\end{subfigure}
\begin{subfigure}{0.5\textwidth}
    \vspace{1em}
    \centering
    \includegraphics[scale=0.35]{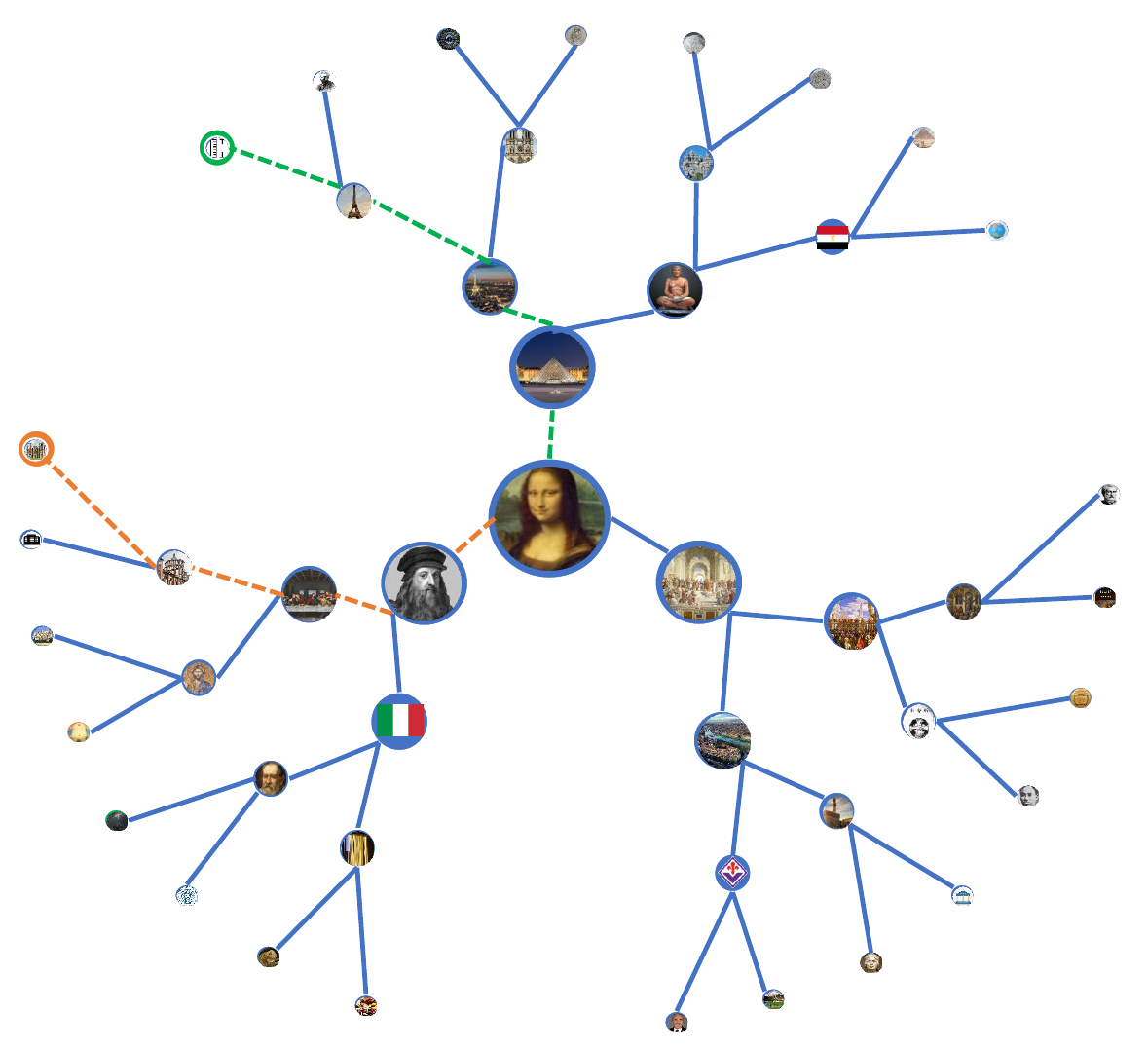}
    \caption{The same hierarchy instantiation is shown in a Lorentz space with negative curvature.}
    \label{Hyper}
\end{subfigure}
\caption{Demonstration of instantiated hierarchies in Euclidean and hyperbolic space. The entity \textit{330m} denoted by the green circle represents the height of \textit{Eiffel Tower} and \textit{Gothic architecture}, highlighted by the orange circle, signifies the architecture style of \textit{Santa Maria delle Grazie}. These two entities, despite being the $4$-hop neighbors of \textit{Mona Lisa}, bear little relevance to each other. Therefore, they should be depicted as somewhat distanced on the graph to reflect their weak romantic correlation.}
\label{Euclidea}
\end{figure}
\textit{Opportunity: Why do we need to represent in hyperbolic space?}
As depicted in Figure~\ref{Euclidea}, we aim to highlight the advantage of using hyperbolic space to represent hierarchical structures by instantiating $\textit{Mona Lisa}$ and expanding its related entities. In the Euclidean space (Figure~\ref{Eu}), the distance between green nodes \textit{330m} and orange nodes \textit{Gothic architecture} is $8$ nodes, as indicated by the orange and green dashed lines, It's observed that embedding tree-like structures into Euclidean space places them in close proximity~\cite{kennedy2013hyperbolicity,adcock2013tree}, but they lack semantics relevance. 

However, in Lorentz space (Figure~\ref{Hyper}), characterized by negative curvature, nodes such as \textit{330m} and \textit{Gothic architecture} near the edge of hyperbolic space may exhibit an actual increase in distance due to spatial curvature~\cite{DBLP:journals/tbd/ZhangWSJY22,sala2018representation,law2019lorentzian}, aligning more closely with the reality.
Therefore, a strong correlation exists between tree-like graphs and hyperbolic space. The aforementioned expansion motivates us to consider representation learning in hyperbolic space.
Additionally, the mapping between the Lorentz space $\Hbb^n_k$ and its tangent space (Euclidean space) $\Gamma_x\Hbb^n_k$ can be established using the exponential map and logarithmic map. The definitions are as follows:
\begin{align*}
     & \Gamma_x \Hbb^n_k \rightarrow \Hbb^n_k: \exp_\xv^k (\vv) = \cosh (\alpha)\xv + \sinh(\alpha) \frac{\vv}{\alpha},            
      \alpha = \sqrt{|k|} \| \vv \|_\Hbb,
    \| \vv \|_\Hbb = \sqrt{\langle\vv,\vv \rangle_\Hbb};   \\
     & \Hbb^n_k \rightarrow \Gamma_x\Hbb^n_k :\log_\xv^{k}(\yv) = \frac{\cosh^{-1}\beta}{\sqrt{\beta^{2}-1}} (\yv-\beta \xv), 
     \beta = k \langle \xv,\yv \rangle_\Hbb
\end{align*}

\begin{figure*}[ht]
\centering
    \subfloat[Visualization of dimensionality reduced embeddings, obtained from the UniGNN model in Euclidean space, using the UMAP technique.]{\includegraphics[scale=0.95]{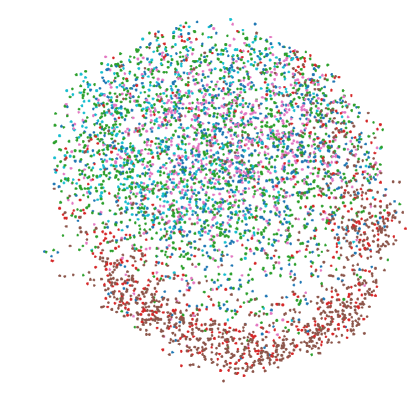}}
    \hspace{1cm}
    \subfloat[Visualization of dimensionality reduced embeddings, obtained from the UniGNN model in Lorentz space, using the UMAP technique.]{\includegraphics[scale=0.95]{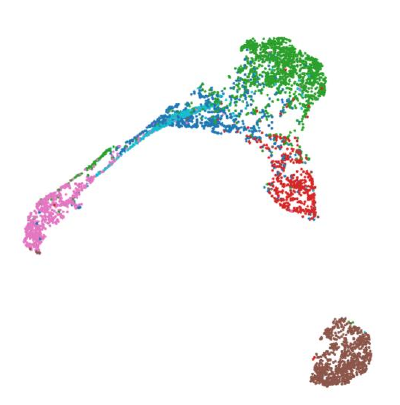}}
    \vspace{1em}
    \caption{Visualization $2$-dimensional representations on the DBLP dataset in both Euclidean and Lorentz spaces. Each point denotes an individual author, with color differentiation illustrating their respective labels.}
\label{practice}
\end{figure*}
In practice, we employ the DBLP dataset as a case to visualize the embeddings acquired through UniGNN~\cite{DBLP:conf/ijcai/HuangY21}, a unified framework for graph and hypergraph neural networks. 
We apply the uniform manifold approximation and projection (UMAP) technique~\cite{DBLP:journals/corr/abs-1802-03426} for two-dimensional dimensionality reduction, representing the embedding in both Euclidean and Lorentz spaces. As depicted in Figure~\ref{practice}, each point in the visualization corresponds to an author, with color indicating the respective label.
Notably, the Lorentz space relatively clearly separates six classes of nodes, exhibiting a clearer boundary and larger discrimination.

\section{Proposed Method}
\begin{figure*}[h]
    \centering
    \includegraphics[scale = 0.23]{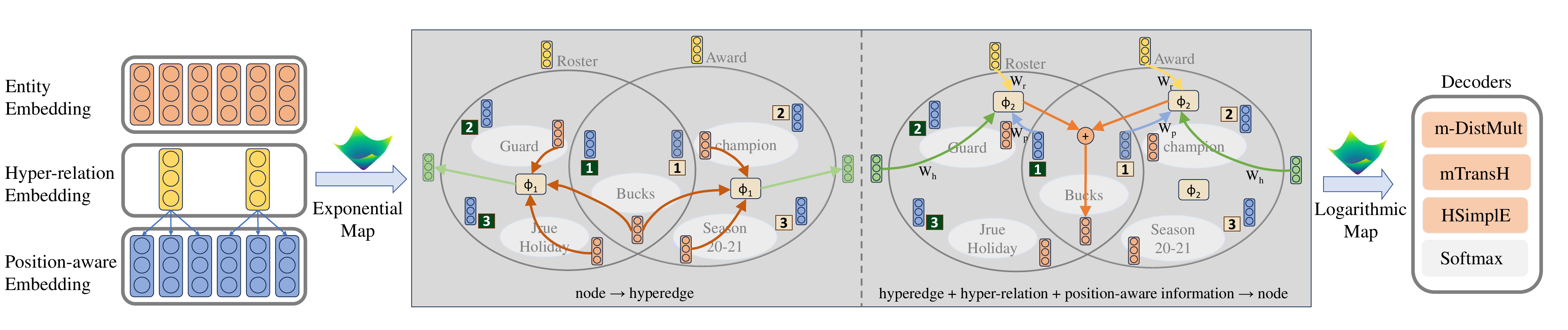}
    \caption{Illustration showcasing the H$^2$GNN architecture with operations executed in Lorentz space through the \textit{Exponential Map}. The two-stage hyper-star message passing, denoted as $\phi_1$ and $\phi_2$, is intricately designed for knowledge hypergraphs. In the function $\phi_1$, the aggregation operation refines hyperedge embeddings, exemplified through the updating process of hyperedges like \texttt{Roster} and \texttt{Award}. Transitioning to $\phi_2$, node embeddings undergo modification by integrating aggregated hyperedge embeddings, hyper-relation embeddings, and position-aware information. Before $\phi_2$, these components individually experience a linear transformation as demonstrated by the matrices $\textbf{W}_h$, $\textbf{W}_r$, and $\textbf{W}_p$. A specific instantiation of the node update process is demonstrated for the node \textit{Bucks}. The decoder incorporates m-DistMult, mTransH, and HsimplE for link prediction, while Softmax is applied to node classification tasks.
    }
    \label{fig:framework}
\end{figure*}

\noindent
In knowledge hypergraphs, each tuple $(r,x_1,x_2, \dots, x_m)$ represents a knowledge fact, where $x_1,x_2, \dots, x_m$ denote the entities, and $r$ represents the hyper-relation, where $m$ is called the arity of hyper-relation $r$. We convert the knowledge tuples to hypergraph $\Gcal = (\Vcal, \Rcal, \Ecal)$, where $\mathcal{V}$ denotes the set of nodes, $\Ecal$ is the set of hyperedges, $\mathcal{R}$ denotes the set of hyper-relations.
The goal of representation learning is to obtain embedded representations for each node $x \in \mathcal{V}$ and hyper-relation $r \in \mathcal{R} $ in the knowledge hypergraphs. 
In this section, we provide a detailed description of H$^2$GNN, and the overall framework is illustrated in Figure~\ref{fig:framework}.

\subsection{Hyperbolic Operations}
\noindent
We denote a hyperboloid space $\Hbb^n$ with negative curvature $k$. In this space, each point is represented as $\xv = \begin{bmatrix} x_t; \xv_s \end{bmatrix}$, where the vector in an $n$-dimensional hyperbolic space is $n+1$ dimensional. 
Here, $x_t$ represents the time dimension, while the remaining $n$ dimensions, denoted as $\xv_s$, represent space.
$\langle \xv, \yv \rangle_\Hbb = -x_ty_t + \xv_s^{\top} \yv_s$.
Both linear and nonlinear operations in hypergraph neural networks within the hyperbolic space should ensure they preserve the space's integrity.
Therefore, we introduce two Lorentz transformations in our work.

\textit{Linear transformation} 
learns a matrix $\textbf{M} = \begin{bmatrix} \vv^\top; \textbf{W} \end{bmatrix}$, where $\vv \in \Rbb^{n+1}$, $\textbf{W} \in \Rbb^{m \times (n + 1)}$, satisfying $\Fcal_x(\textbf {M}) \xv \in \Hbb^m$ and $\Fcal_x (\textbf{M}): \Rbb ^{(m+1) \times (n+1)} \rightarrow \Rbb^{(m+1) \times (n+1)}$ transforms any matrix into an appropriate value for the hyperbolic linear layer. 
Following the~\cite{chen2021fully}, the implementation of the $\Fcal_x (\textbf{M})\xv$ can be:
\begin{align}
    \Fcal_x (\textbf{M}) \xv =  \begin{bmatrix}  \sqrt{\| \phi(\textbf{W}\xv, \vv) \|^2 - 1 / k)} ;\phi(\textbf{W}\xv, \vv) \end{bmatrix}^\top\,,
\end{align}
where $\xv \in \Hbb^n$, $\mathbf{W} \in \Rbb^{m \times (n+1)}$, and $\phi$ is an operation function. 
For activation and normalization, the function can be written as:
\begin{align}
    \phi(\mathbf{W}\xv,\vv) = \frac{\lambda \sigma(\vv^\top \xv+b')}{\|\mathbf{W}h(\xv)+\bv \|} (\mathbf{W}h(\xv)+\bv)\,,
\end{align}
where $\sigma$ represents the sigmoid function, $\bv$ and $b'$ are biases, $\lambda > 0$ controls scaling range, and $h$ denotes the activation function. 
The instantiated linear transformation guarantees the outputs remain in the hyperbolic space, a detailed proof and derivation process can be found in~\cite{chen2021fully}.

\textit{Aggregation operation} enhances to capture the complex relations in graph-structured data. 
we employ the aggregation operation operation using the weighted aggregation squared Lorentzian distance w.r.t $centroid$. 
This operation estimates the statistical dispersion of a set of points~\cite{law2019lorentzian,chen2021fully}. 
Specifically, for a point set $P = \{\yv_1, \dots, \yv_{|P|}\}$, the centroid whose expected distance to $P$ is minimum, i.e. $argmin_{\omegav \in \Hbb_k^n} \sum_{i = 1}^{|P|} v_i d_H^2(\yv_i,\omegav)$, where $v_i$ is the weight of the $i$-th point, $d_H^2(a,b) = 2/k - 2\langle a, b\rangle_H$. 
In our method, we simplify the process and set the weight $\{ v_1, \dots, v_{|P|} \}$ as all-ones set, and the squared Lorentzian distance can be formulated as
\begin{align*}
    &centroid(\{ v_1, \dots, v_{|P|} \}, \{\yv_1, \dots, \yv_{|P|}\}) = \\
    &\frac{\sum_{j=1}^{|P|} v_j \yv_j}{\sqrt{-k} |\| \sum_{i= 1}^{|P|} v_i \yv_i \|_{\Hbb}|}  = \frac{\sum_{j=1}^{|P|}\yv_j}{\sqrt{-k} |\| \sum_{i= 1}^{|P|} \yv_i \|_{\Hbb}|}\,,
\end{align*}
where $k$ is negative curvature, 
$|\|a\|_{\Hbb}|$ is defined as the Lorentzian norm of $a$, $|\|a\|_{\Hbb}| = \sqrt{|\|a\|_{\Hbb}^2|}$, $\| a \|_{\Hbb}^2 = \langle a, a \rangle_\Hbb$ is the squared Lorentzian norm of $a$.

\subsection{Hyper-Star Message Passing}
\noindent
General GNNs leverage the feature matrix and graph structure to obtain informative embeddings for a given graph. The node embeddings undergo iterative updates by incorporating information from their neighboring nodes. 
The message-passing process in the $l$-th layer of GNN is formulated as follows:
\begin{align*}
    \xv_i^{l+1} = \phi^{l} (\xv_i^{l}, \{\xv_j^{l}\}_{j \in \Ncal_{i}}) \,,
\end{align*}
where $\Ncal_i$ denotes the collection of neighboring nodes of node $\xv_i$. $\phi^l$ defines the aggregation operation of the $l$-th layer. The message-passing process in homogeneous hypergraphs can be succinctly expressed as follows:
\begin{align}
\label{message-passing-1}
    \begin{cases}
        \hv_e = \phi_1(\{\xv_j\}_{j \in e})\,, \\
        \xv_i = \phi_2(\xv_i,\{\hv_e\}_{e \in \Ecal_i})\,,
    \end{cases}
\end{align}
where $\Ecal_i$ represents the set of all hyperedges containing the node $\xv_i$. 
The equation utilizes permutation-invariant function $\phi_1$ and GNN aggregation operation $\phi_2$ to aggregate messages from nodes and hyperedges respectively~\cite{DBLP:conf/ijcai/HuangY21}.

However, the existing message-passing frameworks are insufficient for handling the challenges posed by knowledge hypergraphs, requiring consideration of two additional aspects.
$1)$ Knowledge hypergraphs frequently encompass diverse hyper-relation types, necessitating integrating different relation-type information into the message-passing process. $2)$ Hyperedges are typically represented by tuples (i.e., $(r,e_1,e_2, \dots, e_m)$), where the order of entities in an $m$-tuple indicates their role in the hyper-relation, akin to subject and object roles in simple graphs. 
Therefore, position-aware information regarding entities participating in hyper-relations is also crucial during the message-passing process.

We introduce a $d$-dimensional position-aware feature, denoted as $\hv_p \in \Rbb^{d}$, which is specialized by relation type. That implies that hyperedges of the same relation type share identical position-aware representations. 
To seamlessly integrate the position-aware embedding into the message passing framework, we employ composition operations. This integration is expressed mathematically through the following equation:
\begin{align}
    \begin{cases}
        \hv_e = \phi_{1} (\{\xv_j\}_{j \in e}) \\
        \xv_i = \phi_{2}({\xv_i, comp({\hv_e, \rv_e, \hv_p)_{e \in \Ecal_i}}})\,,
    \end{cases}
\end{align}
where $\hv_p$ varies across distinct relation types and different positions. $comp$ represents a composition operation employed to seamlessly integrate the position-aware embedding $\hv_p$, hyperedge embedding $\hv_e$, and relation-based embedding $\rv_e$ during the message aggregation process.

Inspired by~\cite{chen2021fully}, we evaluate one non-parametric simple operator in Lorentz space, defined as:
$ comp(\hv_e, \rv, \hv_p) = centroid (\hv_e, \rv, \hv_p) $, which uses the centroid features as the aggregation result. Moreover, $\phi_1$ and $\phi_{2}$ are implemented through aggregation operations $centroid$, ensuring that these operations remain within the hyperbolic space. 
Section~\ref{experiments} demonstrates the effective performance of our encoder based on the simple design.
Therefore, we present a straightforward yet highly effective instantiation framework, which considers the valuable neighborhood information:
\begin{align}
    \begin{cases}
        \hv_e = centroid (\{\xv_j\}_{j \in e})\,, \\
        \xv_i = centroid(\xv_i, \hv_e, \rv_e, \hv_p)_{e \in \Ecal_i})\,,
    \end{cases}
\end{align}

Given the complexity of hyperbolic geometry computation, we purposefully select a non-parametric simple operator, the $centroid$~\cite{law2019lorentzian}, as the aggregation result. Besides, unlike the typical hybrid approach, where features oscillate between hyperbolic space and tangent space via mapping functions, our operations are fully within hyperbolic space, further reducing time complexity.

\subsection{Objective Function and Optimization}
\noindent
For the node classification task, we use the negative log-likelihood loss to optimize our model by minimizing the difference between the predicted log probabilities and the ground truth labels of each node. 
\begin{equation}
    \mathcal{L} = -\sum_{i=1}^{N}\sum_{j=1}^{C} l_{\{y_i = j\}} \log P(y_i = j | \xv_i)\,,
\end{equation}
where $N$ represents the number of entities, $C$ represents the number of labels. The indicator function $l_{\{y_i = j\}}$ takes the value of $1$ when the gold label $y_i$ equals $j$, and $0$ otherwise. $P(y_i = j | \xv_i)$ represents the model's predicted probability that node $i$ belongs to label $j$.

We train our model on positive and negative instances for the link prediction task. 
Negative samples are generated employing a methodology akin to HypE ~\cite{fatemi2019knowledge}. 
Specifically, we create $N*r$ negative samples for every positive sample from the dataset by randomly replacing each correct entity with $N$ other entities. Here, $N$ serves as a hyperparameter and $r$ denotes the number of entities in the tuple.
For any tuple $\xv$ in training dataset $\Ecal_{train}$, $neg(\xv)$ is utilized to generate a set of negative samples, following the process above. To compute our loss function, we define the cross-entropy loss as follows:
\begin{equation}
    \Lcal = \sum_{\xv \in \Ecal_{train}} -\log \frac{\exp{g(\xv)}}{\exp{g(\xv)} + \sum_{\xv' \in neg(\xv)}\exp{g(\xv')}}
\end{equation}
where $g(\xv)$ signifies our model predicts the confidence score for the tuple $\xv$.

\section{Experiments}
\noindent
\label{experiments}
In this section, we evaluate H$^2$GNN in transductive learning and inductive learning for node classification and link prediction tasks.
Given a hypergraph $\mathcal{G}$, consisting of node data $\mathcal{V}$ and hyperedges $\mathcal{E}$, the node classification and inductive learning tasks involve developing a classification function that assigns labels to nodes. The link prediction task focuses on predicting new links between entities within the hypergraph, leveraging the existing connections as a basis. 

\begin{table*}[h]
    \caption{Statistics on the dataset, `classes' and `relations' are the number of node types and hyperedge types, respectively.}
    \vspace{1em}
    \centering
    \label{data1}
    \renewcommand{\arraystretch}{1.1}
    \scalebox{1.0}{
        \tabcolsep = 0.07cm
        \begin{tabular}{lccccccc}
\toprule
               & \begin{tabular}[c]{@{}c@{}}DBLP\\ (Co-authorship)\end{tabular} & \begin{tabular}[c]{@{}c@{}}Cora\\ (Co-authorship)\end{tabular} & \begin{tabular}[c]{@{}c@{}}Cora\\ (Co-citation)\end{tabular} & \begin{tabular}[c]{@{}c@{}}Pubmed\\ (Co-citation)\end{tabular} & \begin{tabular}[c]{@{}c@{}}Citeseer\\ (Co-citation)\end{tabular} & \begin{tabular}[c]{@{}c@{}}JF17K\\ (Knowledge Base)\end{tabular} & \begin{tabular}[c]{@{}c@{}}FB-AUTO\\ (Knowledge Base)\end{tabular} \\ \cline{2-8} 
entities & 43,413 & 2,708  & 2,708  & 19,717  & 3,312   & 29,177   & 3,388  \\
hyperedges & 22,535 & 1,072  & 1,579  & 7,963  & 1,079 & 102,648 & 11,213 \\
classes & 6 & 7  & 7  & 3  & 6   & -    & -    \\
relations      & -    & -    & -  & -     & -  & 327    & 8    \\
\#2-ary & 9,976   & 486 & 623 & 3,522   & 541  & 56,332  & 3,786  \\
\#3-ary  & 4,339   & 205  & 464  & 1,626  & 254   & 34,550  & 0    \\
\#4-ary & 2,312  & 106   & 312 & 845   & 118    & 9,509    & 215   \\
\#5-ary   & 1,419  & 78 & 180 & 534  & 65  & 2,230   & 7,212   \\
\#6-ary & 906 & 45  & 0  & 297 & 40  & 37   & 0   \\ \bottomrule
\end{tabular}                                                       }
    \label{dataset}
    \vspace{-1em}
\end{table*}

\begin{table*}[h]
    \centering
    \caption{The accuracy(\%) of node classification on co-authorship and co-citation datasets for baseline methods and H$^2$GNN. The most competitive results are highlighted for each dataset.}
    \vspace{1em}
    \renewcommand{\arraystretch}{1.2}
    \scalebox{1.0}{
        \tabcolsep = 0.2cm
        \begin{tabular}{lccccc}
            \toprule
            \multirow{2}{*}{\textbf{Method}}        & \multicolumn{2}{c}{\textbf{Co-authorship Data}} & \multicolumn{3}{c}{\textbf{Co-citation Data}}  \\ \cline{2-6} & DBLP  & Cora & Cora & Pubmed & Citeseer \\ \hline
            UniSAGE~\cite{DBLP:conf/ijcai/HuangY21} & 88.29$\pm$0.22                                  & 74.04$\pm$1.50                                & 67.08$\pm$2.32            & 74.34$\pm$1.56            & 61.27$\pm$1.78             \\
            UniGIN~\cite{DBLP:conf/ijcai/HuangY21}  & 88.34$\pm$0.21                                  & 73.82$\pm$1.36                                & 66.94$\pm$2.07            & 74.46$\pm$1.81            & 61.09$\pm$1.60             \\
            HyperSAGE~\cite{arya2020hypersage}     & 77.25$\pm$3.11                                  & 72.21$\pm$1.40                                & 66.84$\pm$2.27            & 72.33$\pm$1.18            & 61.08$\pm$1.72             \\
            HyperGCN~\cite{yadati2019hypergcn}      & 71.17$\pm$8.73                                  & 63.29$\pm$7.11                                & 62.43$\pm$9.17            & 67.91$\pm$9.43            & 57.98$\pm$7.01             \\
            FastHyperGCN~\cite{yadati2019hypergcn} & 67.86$\pm$9.46                                  & 61.60$\pm$7.99                                & 61.42$\pm$10.03           & 65.17$\pm$10.03           & 56.76$\pm$8.10             \\
            HGNN~\cite{feng2019hypergraph}          & 68.08+5.10                                      & 63.21$\pm$3.02                                & 68.01$\pm$1.89            & 66.45$\pm$3.17            & 56.99$\pm$3.43             \\  \hline
            \textbf{H$^2$GNN (Ours)}                 & {\textbf{89.75$\pm$0.20}}                       &  \textbf{74.97$\pm$1.20}                    & \textbf{69.43$\pm$1.54} & \textbf{74.89$\pm$1.23} &  \textbf{62.52$\pm$1.48} \\ \bottomrule
        \end{tabular}
    }
    \label{node classification}
\end{table*}

\begin{table*}[h]
    \centering
    \caption{The accuracy(\%) results for inductive learning on evolving hypergraphs across Co-authorship and Co-citation datasets. We highlight the best and most competitive results achieved by baselines and H$^2$GNN for each dataset.}
    \vspace{1em}
    \renewcommand{\arraystretch}{1.2}
    \scalebox{1.0}{
        \tabcolsep = 0.3cm
        \begin{tabular}{lcccccccc}
            \toprule
            \multirow{2}{*}{\textbf{Method}}        & \multicolumn{2}{c}{DBLP} & \multicolumn{2}{c}{Pubmed} & \multicolumn{2}{c}{Citeseer} & \multicolumn{2}{c}{Cora (Co-citation)}       \\ \cline{2-9}
& seen                     & unseen                     & seen                         & unseen                                & seen                     & unseen                  & seen                     & unseen                  \\ \hline
            UniGIN~\cite{DBLP:conf/ijcai/HuangY21}  & 89.4$\pm$0.1             & 83.2$\pm$0.2               & 84.5$\pm$0.3                 & 83.1$\pm$0.4                          & 69.1$\pm$1.1             & 68.8$\pm$1.7            & 71.6$\pm$2.0             & 68.7$\pm$2.1            \\
            UniSAGE~\cite{DBLP:conf/ijcai/HuangY21} & 89.3$\pm$0.2             & 82.7$\pm$0.3               & 80.3$\pm$1.0                 & 79.2$\pm$0.8                          & 67.9$\pm$1.5             & 68.2$\pm$1.2            & 70.5$\pm$1.1             & 66.3$\pm$1.4            \\
            UniGCN~\cite{DBLP:conf/ijcai/HuangY21}  & 88.1$\pm$0.2             & 82.1$\pm$0.1               & 17.6$\pm$0.3                 & 17.8$\pm$0.3                          & 22.1$\pm$0.8             & 22.4$\pm$0.8            & 15.6$\pm$0.9             & 15.8$\pm$0.9            \\
            UniGAT~\cite{DBLP:conf/ijcai/HuangY21}& 88.0$\pm$0.1             & 15.8$\pm$0.2               & 30.0$\pm$0.4                 & 17.8$\pm$0.2                          & 44.2$\pm$0.6             & 22.5$\pm$0.6            & 48.3$\pm$1.0             & 15.8$\pm$0.5            \\ \hline
            \textbf{H$^2$GNN (Ours)}                 &  \textbf{89.7$\pm$0.1} &  \textbf{83.4$\pm$0.2}   &  \textbf{86.2$\pm$0.2}     &  \textbf{85.5$\pm$0.5}              &  \textbf{70.2$\pm$1.3} & \textbf{69.2$\pm$1.0} &  \textbf{75.3$\pm$1.4} & \textbf{72.1$\pm$1.2} \\ \bottomrule
        \end{tabular}
    }
    
    \label{inductive}
\end{table*}

\begin{table*}[h]
    \centering
    \caption{Knowledge Hypergraph link prediction results on JF17k and FB-AUTO for baselines and H$^2$GNN. The G-MPNN method did not produce results on the JF17k dataset for two days, so the experimental results are not shown.}
    \vspace{1em}
    \renewcommand{\arraystretch}{1.1}
   \scalebox{1.0}{
       \begin{tabular}{lcccccccc}
\toprule & \multicolumn{4}{c}{FB-AUTO}    & \multicolumn{4}{c}{JF17K} \\ \cline{2-9} \multirow{-2}{*}{Method}                                                           & Hits@1                       & Hits@3                     & Hits@10          & \multicolumn{1}{c|}{MRR}             & Hits@1           & Hits@3           & Hits@10          & MRR             \\  \hline
            \multicolumn{1}{l|}{{\color[HTML]{333333} m-TransH~\cite{wen2016representation}}}        & 0.602                      & 0.754                    & 0.806          & \multicolumn{1}{c|}{0.688}          & 0.370            & 0.475           & 0.581           & 0.444           \\
            \multicolumn{1}{l|}{{\color[HTML]{333333} m-CP~\cite{fatemi2019knowledge}}}              & 0.484                       & 0.703                     & 0.816           & \multicolumn{1}{c|}{0.603}          & 0.298           & 0.443           & 0.563           & 0.391           \\
            \multicolumn{1}{l|}{{\color[HTML]{333333} m-DistMult~\cite{fatemi2019knowledge}}}         & 0.513                      & 0.733                    & 0.827          & \multicolumn{1}{l|}{0.634}          & 0.372           & 0.510            & 0.634           & 0.463           \\
            \multicolumn{1}{l|}{{\color[HTML]{333333} r-SimplE~\cite{fatemi2019knowledge}}}& 0.082                       & 0.115                     & 0.147           & \multicolumn{1}{l|}{0.106}           & 0.069           & 0.112           & 0.168           & 0.102           \\
            \multicolumn{1}{l|}{{\color[HTML]{333333} NeuInfer~\cite{DBLP:conf/acl/GuanJGWC20}}}             & \textbf{0.700}                         & 0.755                     & 0.805           & \multicolumn{1}{c|}{0.737}           & 0.373           & 0.484           & 0.604           & 0.451           \\
            \multicolumn{1}{l|}{{\color[HTML]{333333} HINGE~\cite{rosso2020beyond}}}                 & 0.630                        & 0.706                     & 0.765           & \multicolumn{1}{l|}{0.678}           & 0.397           & 0.490            & 0.618           & 0.473           \\
            \multicolumn{1}{l|}{{\color[HTML]{333333} NALP~\cite{guan2019link}}}                     & 0.611                       & 0.712                     & 0.774           & \multicolumn{1}{l|}{0.672}           & 0.239           & 0.334           & 0.450            & 0.310            \\
            \multicolumn{1}{l|}{{\color[HTML]{333333} RAE~\cite{zhang2018scalable}}}                 & 0.614                       & 0.764                     & 0.854           & \multicolumn{1}{l|}{0.703}           & 0.312           & 0.433           & 0.561           & 0.396           \\
            \multicolumn{1}{l|}{{\color[HTML]{333333} HypE~\cite{fatemi2019knowledge}}}              & 0.662            & 0.800                    & 0.844          & \multicolumn{1}{l|}{0.737}          & 0.403 & 0.531           & 0.652          & 0.489
            \\
            \multicolumn{1}{l|}{{\color[HTML]{333333}
            G-MPNN~\cite{DBLP:conf/nips/Yadati20}}}                                                  & 0.201                       & 0.407                    & 0.611          & \multicolumn{1}{l|}{0.337}          & -               & -               & -               & -
            \\ 

            \multicolumn{1}{l|}{{\textbf{H$^2$GNN (Ours)}}}                                             & 0.674                      & \textbf{0.831}            & \textbf{0.884} & \multicolumn{1}{c|}{{\textbf{0.757}}} & \textbf{0.411}          & \textbf{0.548} & \textbf{0.669} & \textbf{0.498}   \\ \bottomrule
\end{tabular}    
    }
    \label{link prediction}
    \vspace{-1em}
\end{table*}

\subsection{Settings}
\noindent
\textbf{Dataset.}
We employ widely used academic Co-citation and Co-author datasets~\cite {yadati2019hypergcn}, including DBLP, CiteSeer, Pubmed, and Cora for node classification tasks. 
For the link prediction task, our approach is evaluated on two hyper-relation datasets: JF17k~\cite{wen2016representation} and FB-AUTO~\cite{DBLP:conf/sigmod/BollackerEPST08}, which consist of both binary and $n$-ary facts.
Further details and statistics of the datasets can be found in Table~\ref{data1}.

\noindent\textbf{Compared methods.}
For the node classification task, we conduct a comparative analysis between H$^2$GNN and representative baseline methods, including Hypergraph neural networks~\cite{feng2019hypergraph}, HyperGCN~\cite{yadati2019hypergcn}, FastHyperGCN~\cite{yadati2019hypergcn}, HyperSAGE~\cite{arya2020hypersage}, UniGNN ~\cite{DBLP:conf/ijcai/HuangY21}.
In the knowledge hypergraph link prediction, we categorize the introduced baselines into two groups: $(1)$ models that operate with binary relations and can be easily extended to higher-arity: r-SimplE~\cite{DBLP:conf/ijcai/FatemiTV020}, m-DistMult~\cite{fatemi2019knowledge}, m-CP~\cite{fatemi2019knowledge} and m-TransH~\cite{wen2016representation}; and $(2)$ existing methods capable of handling higher-arity relations: NeuInfer~\cite{guan2020neuinfer}, HINGE ~\cite{rosso2020beyond}, NaLP~\cite{guan2019link}, RAE~\cite{zhang2018scalable} and HypE~\cite{fatemi2019knowledge}. 

\noindent
\textbf{Hyper-parameter setting.}
Using PyTorch, we implemented the H$^2$GNN framework and performed the training process on a Tesla V100 GPU machine. The parameters for the other methods are configured according to the recommendations provided by their respective authors.
For the node classification task, we adopt a two-layer H$^2$GNN with the following hyper-parameters: learning rate of $0.01$, weight decay of $5e$$-5$, the dropout rate of $0.5$, and hidden layer dimension of $8$.
We fix the number of training epochs at $200$ and report model performance based on the best validation score on the test dataset for each run.

For the link prediction task, we adopt a single-layer H$^2$GNN with the following hyper-parameters: learning rate of $0.05$, embedding dimension of $200$, the dropout rate of $0.2$, and a negative ratio of $10$. We trained using batches of $128$ items for $2000$ iterations, selecting the model that achieved the highest validation score for testing purposes and recording its results. 
\subsection{Node Classification Results}
\noindent
The semi-supervised node classification task aims to predict labels for test nodes, given the hypergraph structure, node features and limited training labels. Table~\ref{node classification} reveals that H$^2$GNN consistently outperforms other methods across all datasets, demonstrating both high performance and low standard deviation.
This highlights H$^2$GNN's effective representation of hypergraph features. Furthermore, when compared to UniSAGE, which also utilizes a two-stage message-passing schema in the Euclidean space, it becomes evident that hyperbolic space is better suited for modeling hierarchical structural information.

Besides, following the approach~\cite{arya2020hypersage}, we conduct inductive learning on evolving hypergraph, take the historical hypergraphs as input, and predict the unseen nodes' labels. 
Specifically, we use a corrupted hypergraph that randomly removes $40\%$ nodes as unseen data during training. For training, We use $20\%$ of the nodes, reserving $40\%$ for the testing nodes that are seen during training.
H$^2$GNN consistently outperforms other methods across benchmark datasets, achieving a remarkable accuracy of $89.7\%$ on DBLP, with a low standard deviation ranging from $0.1\%$ to $1.4\%$, as shown in Table~\ref{inductive}.
This highlights the effectiveness and robustness of H$^2$GNN for inductive learning on evolving hypergraphs. H$^2$GNN can dynamically update the hypergraph structure and node features, leveraging high-order neighbor information to enhance node representation. 

\subsection{Link Prediction Results}
\noindent
\textit{Knowledge hypergraph completion} can be achieved by either extracting new facts from external sources or predicting links between existing facts in the hypergraph. The latter entails inferring new knowledge from the structure of the hypergraph itself, which is the focus of our experiment~\cite{rossi2021knowledge}. 
We employ H$^2$GNN as the encoder and HSimplE~\cite{DBLP:conf/ijcai/FatemiTV020}  as the decoder, achieving the highest values on Hits@10 evaluation metrics, with scores of $0.884$ on FB-AUTO and $0.669$ on the JF17k dataset, as shown in Table~\ref{link prediction}.
Our method demonstrates a significant improvement over G-MPNN, which was specifically designed for heterogeneous hypergraphs and it also compares favorably with the specialized embedding-based method HypE~\cite{fatemi2019knowledge}, designed for link prediction tasks.

Moreover, we conduct experiments to investigate the impact of different encoders and decoders on the link prediction task. As illustrated in Table~\ref{decoders}, we keep H$^2$GNN as the encoder and pair it with three decoders: HSimplE, mTransH, and m-DistMult. Simultaneously, we examine the effects of using these three decoders individually. The results indicate that the combination of H$^2$GNN-HSimplE, H$^2$GNN-mTransH, and H$^2$GNN-m-DistMult significantly outperform the use of HSimplE, mTransH, and m-DistMult alone. 
This suggests that the encoder-decoder synergy can effectively harness the knowledge hypergraph's structural and semantic information.

\begin{figure*}[ht]
    \centering
    \scalebox{1.0}{
    \includegraphics[width = 0.9\linewidth]{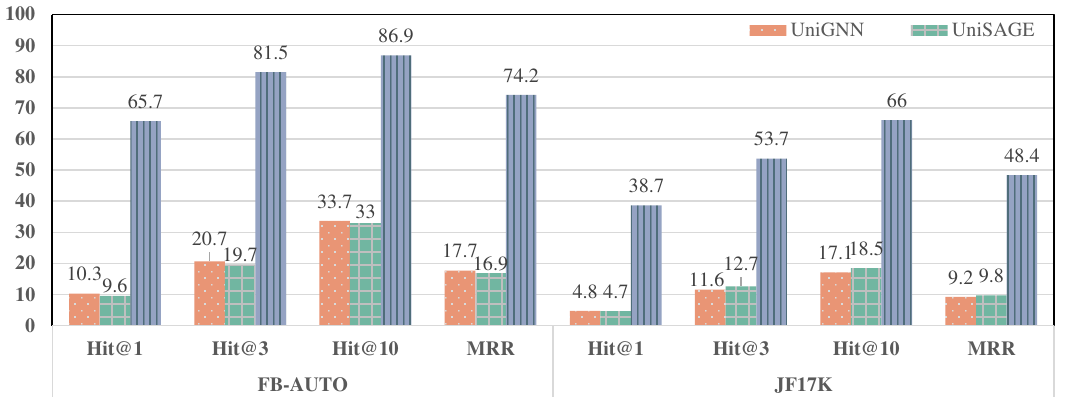}}
    \caption{Comparison experiments: encoding the hypergraph structure information with different methods for the same m-DistMult decoding model.}
    \vspace{-1em}
    \label{encoders}
\end{figure*}

\begin{table*}[!ht]
    \centering
    \caption{Comparison experiments: H$^2$GNN encodes the graph structure information and contrasts experimental results among various decoders. The results are presented in bold. Additionally, the effects of employing each of these three decoders individually are examined. The best results are underlined.}
    \vspace{1em}
    \label{decoders}
    \renewcommand{\arraystretch}{1.1}
    \scalebox{1.0}{
        \begin{tabular}{lllllllll}
            \toprule & \multicolumn{4}{c}{FB-AUTO}               &\multicolumn{4}{c}{JF17K}                       \\ \cline{2-9}
            \multirow{-2}{*}{\textbf{Method}} & {\color[HTML]{000000} Hits@1}             & {\color[HTML]{000000} Hit@3}              & {\color[HTML]{000000} Hits@10}            & {\color[HTML]{000000} MRR}                & {\color[HTML]{000000} Hits@1}             & {\color[HTML]{000000} Hit@3}              & {\color[HTML]{000000} Hits@10}            & {\color[HTML]{000000} MRR}               \\ \hline 
            H$^2$GNN-HSimplE                 &  {\uline{ \textbf{0.674}}} &  {\uline{\textbf{0.831}}} & {\uline{\textbf{0.884}}} &  {\uline{\textbf{0.757}}} & {\uline{\textbf{0.411}}} &  {\uline{\textbf{0.548}}} &  {\uline{\textbf{0.669}}} &  {\uline{\textbf{0.498}}} \\
            HSimplE                          & 0.608                & 0.760                & 0.825                & 0.692                & 0.341                & 0.490               & 0.633                & 0.451               \\
            H$^2$GNN-mTransH                 & \textbf{0.621}                     &  \textbf{0.771}                     & \textbf{0.840}                     &  \textbf{0.705}                     &  \textbf{0.372}                     & \textbf{0.481}                     &  \textbf{0.583}                     &  \textbf{0.451}                    \\
            mTransH                           & 0.602                                    & 0.754                                    & 0.806                                    & 0.688                                    & 0.370                                      & 0.475                                     & 0.581                                     & 0.444                                    \\
            H$^2$GNN-m-DistMult               & \textbf{0.657}                     & \textbf{0.815}                     & \textbf{0.869}                     & \textbf{0.742}                     & \textbf{0.387}                     & \textbf{0.537}                     & \textbf{0.660}                     & \textbf{0.484}                     \\
            m-DistMult                         & 0.513                                    & 0.733                                    & 0.827                                    & 0.634                                    & 0.372                                     & 0.510                                      & 0.634                                     & 0.463                                    \\ \bottomrule
        \end{tabular}
    }
    \vspace{-1em}
\end{table*}

Figure~\ref{encoders} compares UniGNN, UniSAGE, and H$^2$GNN as encoders when paired with the m-DistMult decoder. The experimental results demonstrate that H$^2$GNN-m-DistMult significantly outperforms UniGNN-m-DistMult and UniSAGE-m-DistMult on both datasets, characterized by heterogeneous hypergraphs with multiple relations. 
Furthermore, validating the message-passage schema in Eq.~\ref{message-passing-1}--- utilized by UniGNN and UniSAGE---it becomes evident that this schema is efficient only in homogeneous hypergraphs, falling short in handling the complexity of heterogeneous hypergraphs. 
For instance, on the FB-AUTO knowledge base, featuring $8$ relations, the combination of H$^2$GNN and m-DistMult achieves Hits@1 of $65.7\%$ and MRR of $74.2\%$, while UniGNN and m-DistMult only attain Hits@1 of $10.3\%$ and MRR of $17.7\%$.
\subsection{Ablation Study}
\noindent
\begin{figure*}[h]
    \centering
    \scalebox{1}{
    \includegraphics[width = 1\linewidth]{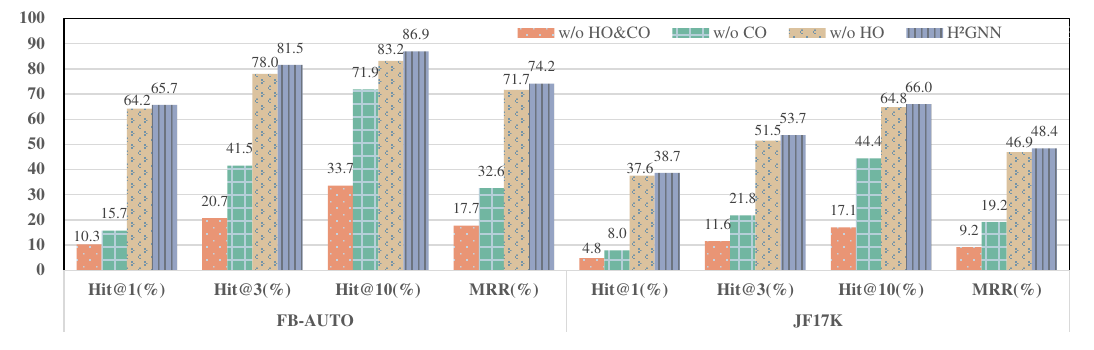}}
    \caption{Sensitivity analysis of H$^2$GNN modules:  hyperbolic operation (HO) and position-aware composition operation (CO) in link prediction task.}
    \vspace{-1em}
    \label{sensitivity}
\end{figure*}

To validate the effectiveness of hyperbolic operation (HO) and position-aware composition operation (CO), we conduct an ablation study on H$^2$GNN.
In this study, we conduct experiments using \textit{m-DistMult} as the decoder, where we respectively remove the two modules and report these model variants' performance in Figure~\ref{sensitivity}. 
The composition operations mainly involve integrating the position-aware features into the massage-passing process. Thus, removing the composition operation implies utilizing Eq.~\ref{message-passing-1} for message passing. 
Additionally, the exclusion of hyperbolic operations entails the implementation of all operations in Euclidean space. In Euclidean space, the averaging operation, which is equivalent to the \textit{centroid} operation in Lorentz space, is employed.
In general, the full model H$^2$GNN exhibits the best performance, and we observe a performance decline when any of the components is removed, highlighting the significance of each component of H$^2$GNN. 
Notably, a substantial decrease in performance is observed after removing the composition operation in terms of Hit@k and MRR. This underscores the importance of position-aware information concerning entities participating in relations, especially in multi-relational knowledge hypergraphs, where capturing semantics and structural information relies heavily on such details. 
One possible explanation is that in simple hypergraphs, relations and tuple schemas are more straightforward and singular. However, in the case of knowledge hypergraphs, the diversity in relation types and the roles of entities in relations introduce complexities, emphasizing the crucial role of position-aware information.

\section{Related work}
\noindent
\noindent\textbf{Graph Neural Networks.} Research in graph neural networks serves as the foundational basis for GNN development.
For instance, Graph Convolutional Networks (GCNs)~\cite{DBLP:conf/iclr/KipfW17} leverage node degrees to normalize neighbor information. 
PPNP~\cite{DBLP:conf/iclr/KlicperaBG19} tackles the over-smoothing problems in GNNs through skip-connections, and AdaGCN~\cite{DBLP:conf/iclr/SunZL21} integrates a traditional boosting method into GNNs.
Heterogeneous graph neural networks~\cite{DBLP:conf/www/HuDWS20,DBLP:journals/tkde/WangSZWY23} have made significant strides in effectively addressing complex heterogeneity through the integration of message passing techniques.
Notably, the heterogeneous graph Propagation Network (HPN)~\cite{DBLP:journals/tkde/JiWSWY23} theoretically provides a theoretical analysis of the deep degradation problem and introduces a convolution layer to mitigate semantic ambiguity.

\noindent\textbf{Hyperbolic Graph Neural Networks.}
Hyperbolic neural networks have demonstrated their ability to effectively model complex data and outperform high-dimensional Euclidean neural networks when using low-dimensional hyperbolic features~\cite{dasgupta2003elementary,giladi2012bourgain}.
While existing hyperbolic networks, such as the hyperbolic graph convolutional neural network~\cite{chami2019hyperbolic}, hyperbolic graph neural network~\cite{liu2019hyperbolic} and multi-relation knowledge graphs like M$^2$GNN~\cite{wang2021mixed}, encode features in hyperbolic space, they are not fully hyperbolic since most of their operations are formulated in the tangent space, which serves as a Euclidean subspace. 
In contrast, fully hyperbolic neural networks, such as FFHR~\cite{shi2023ffhr} define operations that are entirely performed in the hyperbolic space, avoiding the complexities of space operations.

\noindent\textbf{Knowledge Hypergraph Neural Network.} Existing knowledge hypergraph modeling methods are derived from knowledge graph modeling methods, which can be primarily categorized into three groups: translational distance models, semantic matching models, and neural network-based models. 
Translational distance models treat hyper-relations as distances between entities and formulate score functions based on these distances. For instance, models like m-transH~\cite{wen2016representation} and RAE~\cite{zhang2018scalable} generalize the TransH model. They calculate a weighted sum of entity embeddings and produce a score indicating the relevance of the hyper-relation. 
Neural Network-Based Models, like NaLP~\cite{guan2019link} and NeuInfer~\cite{DBLP:conf/acl/GuanJGWC20}, represent hyper-relations using main triples and attribute pairs. They calculate compatibility scores between the main triples and each attribute pair individually using neural networks. The final hyper-relation scores are determined based on these computations.
Semantic Matching Models, such as HypE~\cite{fatemi2019knowledge}, assess the semantic correlation between entities and hyper-relations through matrix products. For instance, HypE incorporates convolution for entity embedding and employs multi-linear products for calculating plausibility scores. HeteHG-VAE~\cite{fan2021heterogeneous} develops a representation mapping between HINs and heterogeneous hypergraphs by introducing identifier node and slave node to model different levels of relations in HINs.
Besides, HeteHG-VAE focuses on knowledge hypergraphs with a single relation type (paper) where node types are explicitly provided. 
H$^2$GNN operates on knowledge hypergraphs with multiple relation and node types, with no mandate for explicit node type specifications within the dataset.

\noindent\textbf{Star Expansion and Clique Expansion in Hypergraphs.} Clique expansion and star expansion are two common expansion strategies, utilized extensively to elevate the conventional representation learning of binary relations to beyond-binary ones~\cite{DBLP:journals/eswa/YangLGWKJ23}. 
Clique expansion~\cite{DBLP:conf/aaai/FengYZJG19,yadati2019hypergcn,DBLP:conf/cikm/YadatiNNYLT20}, in essence, deploys a full connection of pairwise edges to approximate a hyperedge, including all nodes within. 
Although this method facilitates interaction among all nodes of a hypergraph, it fall short in distinguishing between multiple hypergraphs where the fully-connected graph possesses various cuts. Moreover, it overlooks hypergraphs that serve as subgraphs of others.
On the contrary, star expansion, by establishing a bipartite graph whose nodes include hypervertices and hyperedges, morphing a hypergraph into a regular one~\cite{arya2020hypersage}. 
Thus, identical inner nodes with the same hyperedge are interconnected in an indirect manner via the hyperedge node. 
Compared with clique expansion, star expansion retains multiple connections between hypervertices, since every hyperedge is transformed to a node within the expanded graph. Nonetheless, it glosses over the direct connections existing between two hypervertices and two hyperedges. Consequently, information passing from one hypervertex extends to another via a hyperedge node, causing an information decay issue~\cite{DBLP:conf/aaai/YanCWWC24}.

\section{Conclusion}
\noindent
In this study, we depict knowledge facts as hypergraphs and bring forward a hyperbolic hypergraph neural network, denoted as H$^2$GNN, for multi-relational knowledge hypergraph representation. H$^2$GNN pioneers consider the structure-based method and extend GNNs to modeling knowledge hypergraphs, rather than treat the hypergraph as an independent array of $n$-ary facts, without considering the neighborhood information. This unique approach directly encodes adjacent entities, hyper-relations, and entity positions within knowledge hypergraphs based on the innovative hyper-star message-passing schema, considering both structural and positional information. Moreover, we present the hierarchical structure in a Lorentz space.
Our H$^2$GNN encoder yields results comparable to the baselines on knowledge hypergraph link prediction and node classification tasks.

\bibliographystyle{unsrt}  
\bibliography{main}

\end{document}